\pdfoutput=1

\documentclass[11pt]{article}

\usepackage{amsmath,amsfonts,bm}

\def\eqref#1{equation~\ref{#1}}

\def\1{\bm{1}}

\DeclareMathAlphabet{\mathsfit}{\encodingdefault}{\sfdefault}{m}{sl}
\SetMathAlphabet{\mathsfit}{bold}{\encodingdefault}{\sfdefault}{bx}{n}

\usepackage[]{acl}
\usepackage{graphicx} 

\usepackage{times}
\usepackage{latexsym,multirow}

\usepackage{microtype}

\usepackage[T1]{fontenc}

\usepackage{amssymb}
\usepackage{amsmath}

\usepackage[utf8]{inputenc}

\title{Visio-Linguistic Brain Encoding}

\author{Subba Reddy Oota$^{1,2}$, Jashn Arora$^2$, Vijay Rowtula$^2$ \\ \textbf{Manish Gupta$^{2,3}$ and Bapi Raju Surampudi$^2$}\\
$^1$INRIA, Bordeaux, France, $^2$IIIT Hyderabad, India; $^3$Microsoft, India;  \\
\texttt{subba-reddy.oota@inria.fr, \{jashn.arora, vijay.rowtula\}@research.iiit.ac.in}\\
\texttt{gmanish@microsoft.com, raju.bapi@iiit.ac.in}
}

\begin{document}
\maketitle
\begin{abstract}
Enabling effective brain-computer interfaces requires understanding how the human brain encodes stimuli across modalities such as visual, language (or text), etc.
Brain encoding aims at constructing fMRI brain activity given a stimulus. 
There exists a plethora of neural encoding models which study brain encoding for single mode stimuli: visual (pretrained CNNs) or text (pretrained language models). Few recent papers have also obtained separate visual and text representation models and performed late-fusion using simple heuristics. However, previous work has failed to explore: (a) the effectiveness of image Transformer models for encoding visual stimuli, and (b) co-attentive multi-modal modeling for visual and text reasoning. 
In this paper, we systematically explore the efficacy of image Transformers (ViT, DEiT, and BEiT) and multi-modal Transformers (VisualBERT, LXMERT, and CLIP) for brain encoding. 
Extensive experiments on two popular datasets, BOLD5000 and Pereira, provide the following insights. (1) To the best of our knowledge, we are the first to investigate the effectiveness of image and multi-modal Transformers for brain encoding. 
(2) We find that VisualBERT, a multi-modal Transformer, significantly outperforms previously proposed single-mode CNNs, image Transformers as well as other previously proposed multi-modal models, thereby establishing new state-of-the-art. The supremacy of visio-linguistic models raises the question of whether the responses elicited in the visual regions are affected implicitly by linguistic processing even when passively viewing images. Future fMRI tasks can verify this computational insight in an appropriate experimental setting. 



\end{abstract}

\section{Introduction}
In the past decade, artificial neural networks have witnessed a remarkable performance in the computational neuroscience community in understanding how the brain effortlessly performs information perception and processing given various forms of sensory inputs like visual processing in object recognition tasks~\citep{yamins2014performance,cadieu2014deep,eickenberg2017seeing}.
This line of work, namely brain encoding, aims at constructing neural brain activity recordings given an input stimulus. The two most studied forms of stimuli include vision and language.

Since the discovery of the relationship between language/visual stimuli and functions of brain networks~\citep{constable2004sentence,thirion2006inverse}, researchers have been interested in understanding how the neural encoding models predict the fMRI (functional magnetic resonance imaging) brain activity. Recently, several brain encoding models have been developed to (i) understand the ventral stream in biological vision~\citep{yamins2014performance,kietzmann2019recurrence,bao2020map}, and (ii) to study the higher-level cognition like language processing~\citep{gauthier2019linking,schrimpf2020neural,schwartz2019inducing}.
Previous work has mainly focused on independently understanding vision and text stimuli. However, the biological systems perceive the world by simultaneously processing high-dimensional inputs from diverse modalities such as vision, auditory, touch, proprioception, etc.~\citep{jaegle2021perceiver}.
In particular, how the brain effectively processes and provides its visual understanding through natural language and vice versa is still an open question in neuroscience.

There exist a plethora of neural encoding models which predict the brain activity using representations of single-mode stimuli: visual or text.
Convolutional neural networks (CNNs) have been shown to encode semantics from visual stimuli effectively. Interestingly, intermediate layers in deep CNNs trained on ImageNet~\citep{deng2009imagenet} categorization task can partially account for how neurons in intermediate layers of the visual system respond to any given image~\citep{yamins2013hierarchical,yamins2014performance,gucclu2015deep,yamins2016using,wang2019neural}.
However, the more recent and deeper CNNs have not been shown to further improve on measures of brain-likeness, even though their ImageNet performance has vastly increased~\citep{russakovsky2015imagenet}.
Recently,~\cite{kubilius2019brain} proposed a shallow recurrent anatomical network, CORnet, which provided state-of-the-art results on the Brain-score~\citep{schrimpf2020brain} benchmark. Similar to visual encoding models, neural models like deep recurrent neural networks (RNNs), Transformer~\citep{vaswani2017attention} based language models such as BERT~\citep{devlin2019bert}, RoBERTa~\citep{liu2019roberta}, and GPT-2~\citep{radford2019language} have been leveraged to predict the brain activity corresponding to semantic vectors of linguistic items including words, phrases, sentences, and paragraphs~\citep{gauthier2019linking,schrimpf2020neural}. 

However, these pieces of work suffer from the following drawbacks: (1) 
Although these neural encoding models have demonstrated promising results of processing in one of the two brain regions (visual cortex V4 and pre-frontal cortex IT), they still need plenty of efforts to improve on brain encoding for other parts of the brain. Brain encoding for more brain regions is important since input stimuli elicit diverse and distributed representations in the brain, and these activation responses could be repurposed for several novel tasks. (2) They manually choose\footnote{Quoting from~\cite{kubilius2019brain}: ``After testing every layer on both V4 and IT, we report the model's score as the score of the best layer per region.''} particular CNN layers whose activations are used for predicting brain activity specific to the datasets they work with~\cite{kubilius2019brain}; thus, generalization to other datasets is unclear. On the other hand, we observe that using \emph{last layer} activations from VisualBERT leads to best accuracy.

Unlike previous studies, which focus on single-modality (either visual or language stimuli), some authors demonstrated that multi-modal models formed by combining text-based distributional information with visual representations provide a better proxy for human-like intelligence~\citep{anderson2015reading,oota2019stepencog}. 
However, these methods extract representations from each mode separately (image features from CNNs and text features from pretrained embeddings) and then perform a simple late-fusion. Thus, they cannot exploit semantic correspondence across the two modes at different levels effectively. Such late-fusion based multi-modal models are the closest to our work, and our experiments show that our models outperform them.

\begin{figure*}
    \centering
    \includegraphics[width=0.9\textwidth]{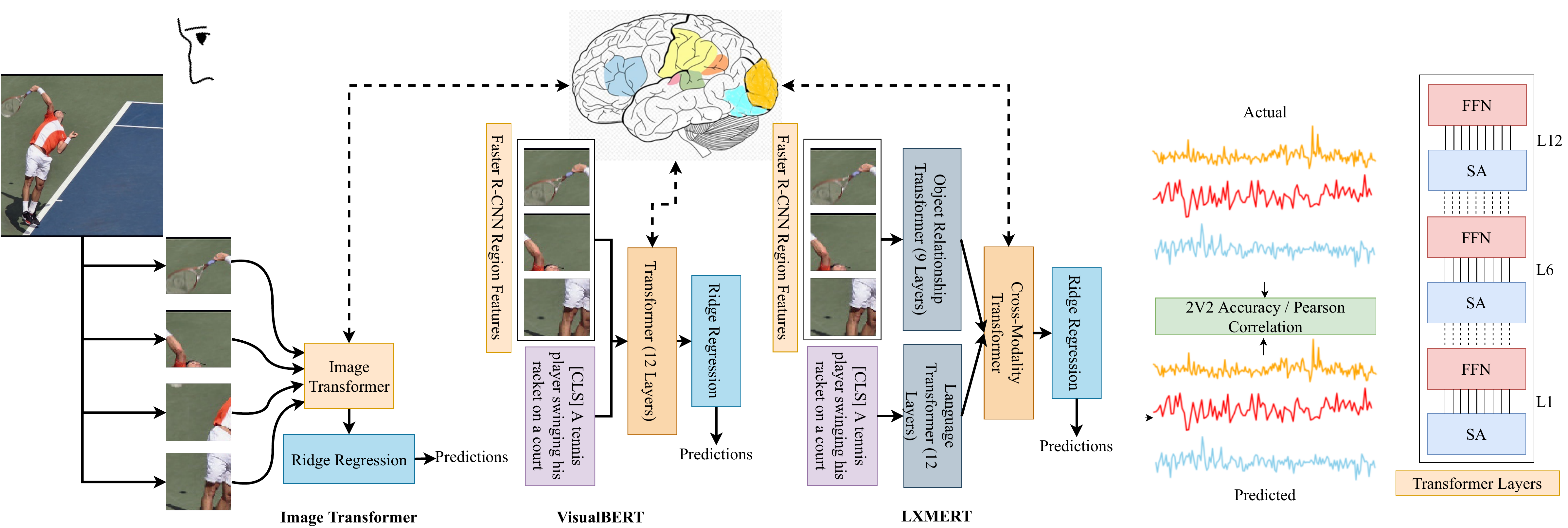}
    \caption{Brain encoding methodology. We use features from image/multi-modal Transformers (like ViT, VisualBERT and LXMERT) as input to the regression model to predict the fMRI activations for different brain regions. Brain encoding results are evaluated by computing 2V2 accuracy and Pearson correlation between actual and predicted activations. We also perform layer-wise correlation analysis between transformer layers and brain regions.}
    \label{fig:arch}
\end{figure*}

Recently, image-based transformer models like ViT~\citep{dosovitskiy2020image}, DEiT~\citep{touvron2021training}, and BEiT~\citep{bao2021beit} have been shown to provide excellent results compared to traditional CNNs on image classification tasks. Also, multi-modal Transformers like VisualBERT~\citep{li2019visualbert}, LXMERT~\citep{tan2019lxmert} and CLIP~\citep{radford2learning} have shown awesome results on visio-linguistic tasks like visual question answering, visual common-sense reasoning, etc.
Inspired by the success of language, image, and multi-modal Transformers, we build multi-modal transformer models to learn the joint representations of image content and natural language and use them for brain encoding. Overall, in this work, we investigate whether \emph{image-based and multi-modal Transformers} can perform fMRI encoding on the \emph{whole brain} accurately. Fig.~\ref{fig:arch} illustrates our brain encoding methodology.


Specifically, we make the following contributions in this paper. 
    (1) We present state-of-the-art encoding results using multi-modal Transformers. We also study the effectiveness of our models in a cross-validated setting.
    (2) Our approach generalizes the use of Transformer-based architectures, removing the need to manually select specific layers as in existing CNN-based fMRI encoding architectures.
    (3) We uncover several cognitive insights about the association between fMRI voxels and representations of multi-modal/image Transformers and CNNs.

\section{Brain Imaging Datasets}
\label{sec:dataset}
The following datasets are popularly used in the literature for studying brain encoding: Vim-1~\citep{kay2008identifying}, Harry Potter~\citep{wehbe2014},~\cite{pereira2018toward}, BOLD5000~\citep{chang2019bold5000}, Algonauts~\citep{cichy2019algonauts} and SS-fMRI~\citep{beliy2019voxels}. Vim-1 has only black and white images and is only related to object recognition, and is subsumed by BOLD5000. SS-fMRI is smaller and very similar to BOLD5000. Harry Potter dataset does not have images. Lastly, fMRIs have not been made publicly available for the Algonauts dataset. Hence, we experiment with BOLD5000 and Pereira datasets in this work. 

\noindent\textbf{BOLD5000}: 
BOLD5000 dataset was collected from four subjects where three subjects viewed 5254 natural images (ImageNet: 2051, COCO: 2135, Scenes: 1068) while fMRIs were acquired. The fourth subject was shown 3108 images only. Details of the visual stimuli and fMRI protocols of the dataset have been discussed in~\citep{chang2019bold5000}. We only briefly summarize the details of the dataset in Table~\ref{tab:bold5000_stats} in the Appendix.
The data covers five visual areas in the human visual cortex, i.e., early visual area (EarlyVis); object-related areas such as the lateral occipital complex (LOC); and scene related areas such as the occipital place area (OPA), the parahippocampal place area (PPA), and the retrosplenial complex (RSC). Each image also has corresponding text labels: ImageNet has a few out of 1000 possible tags per image, COCO has five captions per image, and Scenes has one out of 250 possible categories per image. 

\noindent\textbf{Pereira}: For the Pereira dataset, participants were shown concept word along with a picture with an aim to observe brain activation when participants retrieved relevant meaning using visual information. Sixteen subjects were presented images (six per concept) corresponding to 180 concepts (abstract + concrete), while fMRIs were acquired. Out of 180 concepts, 116 are concrete, and others are abstract. Here, we augmented the image captions using the concept word associated with each image in the picture view.
As in~\citep{pereira2018toward}, we focused on nine brain regions corresponding to four brain networks: Default Mode Network (DMN) (linked to the functionality of semantic processing), Language Network (related to language processing, understanding, word meaning, and sentence comprehension, Task Positive Network (related to attention, salience information), and Visual Network (related to the processing of visual objects, object recognition). We briefly summarize the details of the dataset and the number of voxels corresponding to each region in Table~\ref{tab:pereira_stats} in the Appendix.

\section{Task Descriptions}
For both datasets, we train fMRI encoding models using Ridge regression on stimuli representations obtained using a variety of models as shown in Fig.~\ref{fig:arch}. The main goal of each fMRI encoder model is to predict fMRI voxel values for each brain region given a stimuli. In all cases, we train a model per subject separately. Different brain regions are involved in the processing of stimuli involving objects and scenes. Similarly, some regions specialize in understanding vision inputs while others interpret linguistic stimuli better. To understand the generalizability of our models across these cognitive aspects (objects vs. scenes, language vs. vision), we conduct the following experiments. Whenever we train and test on the same dataset, we follow K-fold (K=10) cross-validation. All the data samples from K-1 folds were used for training, and the model was tested on samples of the left-out fold.

\noindent\textbf{Full dataset fMRI Encoding}: For each dataset, we perform K-fold (K=10) cross-validation. 

\noindent\textbf{Cross-validated fMRI Encoding}: 
In the BOLD5000 dataset, we have three sub-datasets: COCO, ImageNet, and Scenes. For each of the three sub-datasets, we perform K-fold (K=10) cross-validation within the sub-dataset.

ImageNet images mainly contain objects. Scenes images are about natural scenes, while COCO images relate to both objects and scenes. To evaluate the generalizability of our models across objects vs. scenes understanding, we also perform cross-validated experiments where the train images belong to one sub-dataset while the test images belong to the other sub-dataset. Thus, for each subject, we perform (1) three same-sub-dataset train-test experiments and (2) six cross-sub-dataset train-test experiments. 

\noindent\textbf{Abstract vs Concrete Concept fMRI Encoding}: 
Similarly, in the Pereira dataset, we have two sub-datasets: abstract and concrete. Intuitively, concrete images can be interpreted mainly using visual processing, while abstract images may require linguistic processing. Hence, we experiment with two different settings for each subject: (train on abstract, test on concrete) and (train on concrete, test on abstract).



\section{Methodology}
We trained a ridge regression based encoding model to predict the fMRI brain activity associated with the stimuli representation for each brain region. Each voxel value is predicted using a separate ridge regression model.
Formally, we encode the stimuli as $X\in \mathbb{R}^{N \times D}$ and brain region voxels $Y\in \mathbb{R}^{N \times V}$, where $N$ denotes the number of training examples, $D$ denotes the dimension of input stimuli representation, and $V$ denotes the number of voxels in a particular region.

The input stimuli representation can be obtained using any of the following models: (i) pretrained CNNs, (ii) pretrained text Transformers (iii) image Transformers, (iv) late-fusion models, or (v) multi-modal Transformers. The ridge regression objective function for the $i^{th}$ example is given as follows.
\setlength{\belowdisplayskip}{0pt} \setlength{\belowdisplayshortskip}{0pt}
\setlength{\abovedisplayskip}{0pt} \setlength{\abovedisplayshortskip}{0pt}
\begin{align*}
     f(X_{i}) &= \underset{W}{\text{min}} \lVert Y_i - X_{i}W \rVert_{F}^{2} + \lambda \lVert W \rVert_{F}^{2}
\end{align*}
Here, $W$ are the learnable weight parameters, $\lVert.\rVert_{F}$ denotes the Frobenius norm, and $\lambda >0$ is a tunable hyper-parameter representing the regularization weight. $\lambda$ was tuned on a small disjoint validation set obtained from the training part. 

In the following text, we discuss different input stimuli representation methods. Pretrained CNNs and Image Transformers encode image stimuli only, while Pretrained text Transformers encode text stimuli only. Late fusion models and Multi-modal Transformers encode both text and image stimuli. 

\noindent\textbf{Pretrained CNNs}: Inspired by the Algonauts challenge~\citep{cichy2019algonauts}, we extract the layer-wise features from different pretrained CNN models such as VGGNet19~\citep{simonyan2014very} (MaxPool1, MaxPool2, MaxPool3, MaxPool4, MaxPool5, FC6, FC7, FC8), ResNet50~\citep{he2016deep} (Block1, Block2, Block3, Block4, FC), InceptionV2ResNet~\citep{szegedy2017inception} (Conv2D5, Conv2D50, Conv2D100, Conv2D150, Conv2D200, Conv2D\_7b), and EfficientNetB5~\citep{tan2019efficientnet} (Conv2D2, Conv2D8, Conv2D16, Conv2D24, FC), and use them for predicting fMRI brain activity. Here, we use adaptive average pooling on each layer to get feature representation for each image.

\noindent\textbf{Pretrained text Transformers}: RoBERTa~\citep{liu2019roberta} builds on BERT's language masking strategy and has been shown to outperform several other text models on the popular GLUE NLP benchmark. We use the average-pooled representation\footnote{Average-pooled representation gave us better results compared to using the CLS representation.} from RoBERTa to encode text stimuli. 

\noindent\textbf{Image Transformers}: We used three image Transformers: Vision Transformer (ViT), Data Efficient Image Transformer (DEiT), and Bidirectional Encoder representation from Image Transformer (BEiT). Given an image, image Transformers output two representations: pooled and patches. We experiment with both representations. 




\noindent\textbf{Late-fusion models}: In these models, the stimuli representation is obtained as a concatenation of image stimuli encoding obtained from pretrained CNNs and text stimuli encoding obtained from pretrained text Transformers. Thus, we experiment with these late-fusion models: VGGNet19+RoBERTa, ResNet50+RoBERTa, InceptionV2ResNet+RoBERTa and EfficientNetB5+RoBERTa. 
These models do not incorporate real information fusion but just do concatenation across modalities.

\noindent\textbf{Multi-modal Transformers}: We experimented with these multi-modal Transformer models: Contrastive Language-Image Pre-training (CLIP), Learning Cross-Modality Encoder Representations from Transformers (LXMERT), and VisualBERT. These Transformers take both image and text stimuli as input and output a joint visio-linguistic representation. Specifically, the image input for these models comprises of region proposals as well as bounding box regression features extracted from Faster R-CNN~\citep{ren2015faster} as input features as shown in Fig.~\ref{fig:arch}. These models incorporate information fusion across modalities at different levels of processing using co-attention and hence are expected to result in high quality visio-linguistic representations.

\noindent\textbf{Hyper-parameter Settings}: We used sklearn's ridge-regression with default parameters, 10-fold cross-validation, Stochastic-Average-Gradient Descent Optimizer, Huggingface for Transformer models, MSE loss function, and L2-decay ($\lambda$) as 1.0. We used Word-Piece tokenizer for the linguistic Transformer input and Faster-RCNN~\citep{ren2015faster} for extracting region proposals. All experiments were conducted on a machine with 1 NVIDIA GEFORCE-GTX GPU with 16GB GPU RAM.  

\section{Experiments}
\label{sec:experiments}


\subsection{Evaluation Metrics}
\label{sec:metrics}
We evaluate our models using popular brain encoding evaluation metrics described in the following. Given a subject and a brain region, let $N$ be the number of samples. Let $\{Y_i\}_{i=1}^N$ and $\{\hat{Y}_i\}_{i=1}^N$ denote the actual and predicted voxel value vectors for the $i^{th}$ sample. Thus, $Y\in R^{N\times V}$ and $\hat{Y}\in R^{N\times V}$ where $V$ is the number of voxels in that region.

\noindent\textbf{2V2 Accuracy} is computed as follows.
\begin{align*}
    &\text{2V2Acc}=\\
    &\frac{1}{N_{C_2}}\sum_{i=1}^{N-1}\sum_{j=i+1}^N I[\{cosD(Y_i, \hat{Y}_i)+cosD(Y_j, \hat{Y}_j)\}\\
    &<\{cosD(Y_i, \hat{Y}_j)+cosD(Y_j, \hat{Y}_i)\}]
\end{align*}
\noindent where $cosD$ is the cosine distance function. $I[c]$ is an indicator function such that $I[c]=1$ if $c$ is true, else it is 0. The higher the 2V2 accuracy, the better.


\noindent\textbf{Pearson Correlation (PC)} is computed as PC=$\frac{1}{N}\sum_{i=1}^{n} corr[Y_i, \hat{Y}_i]$ where corr is the correlation function.


\begin{figure*}[t]
\centering
\includegraphics[width=0.9\linewidth, height=3.8in]{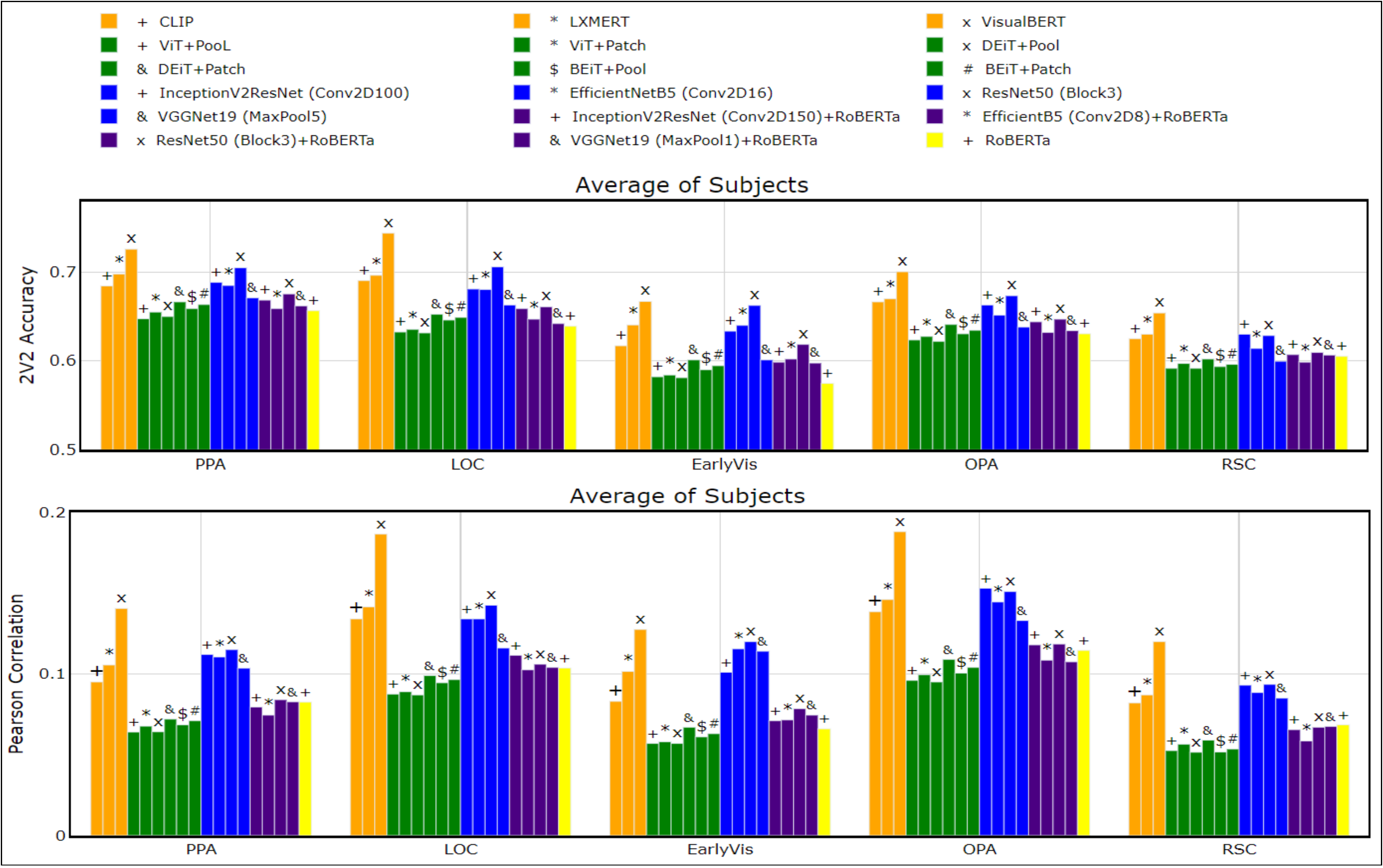}
\caption{BOLD5000 Results: 2V2 (top figure) and Pearson correlation coefficient (bottom figure) between predicted and true responses across different brain regions using a variety of models. Results are averaged across all participants. VisualBERT perform the best.}
\label{fig:Bold_2v2_pcc_avg}
\end{figure*}



\begin{figure}[t]
\centering
 \scriptsize 
\includegraphics[width=1\linewidth]{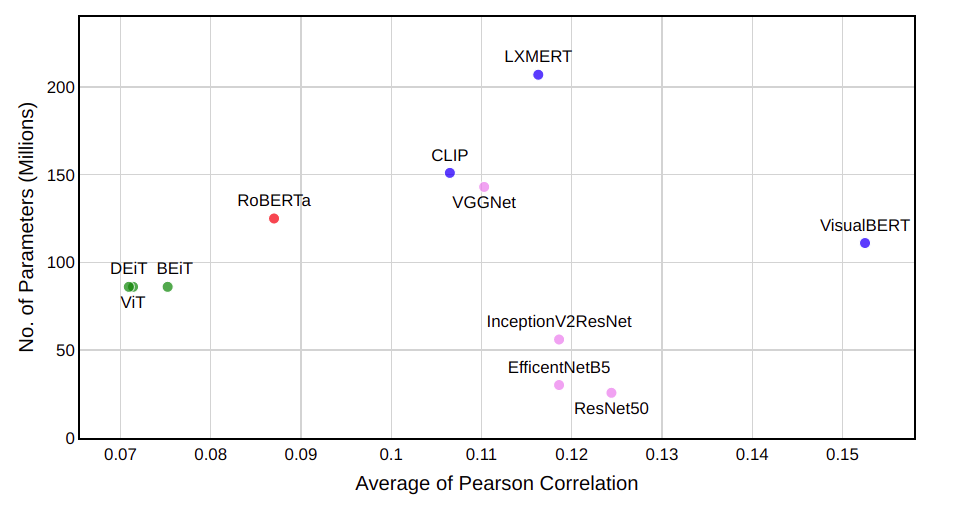}
\caption{BOLD5000: \#Parameters vs Avg Pearson Corr.}
\label{fig:plot1}
\end{figure}


\subsection{Do multi-modal Transformers outperform other models?}

Unfortunately, there is no previous work that uses image Transformers or multi-modal Transformers for brain encoding.
StepEnCog~\citep{oota2019stepencog} is a late-fusion method, but it has a different setting where the model expects voxel values per brain slice rather than per brain region. Besides performing extensive evaluation using a large variety of models, we also compare our results with those obtained by two previously proposed baselines that leverage pretrained CNN models:~\citep{blauch2019assessing} and~\citep{wang2019neural} which use VGGNet. 

\begin{figure*}[t] 
\includegraphics[width=\linewidth, height=3.2in]{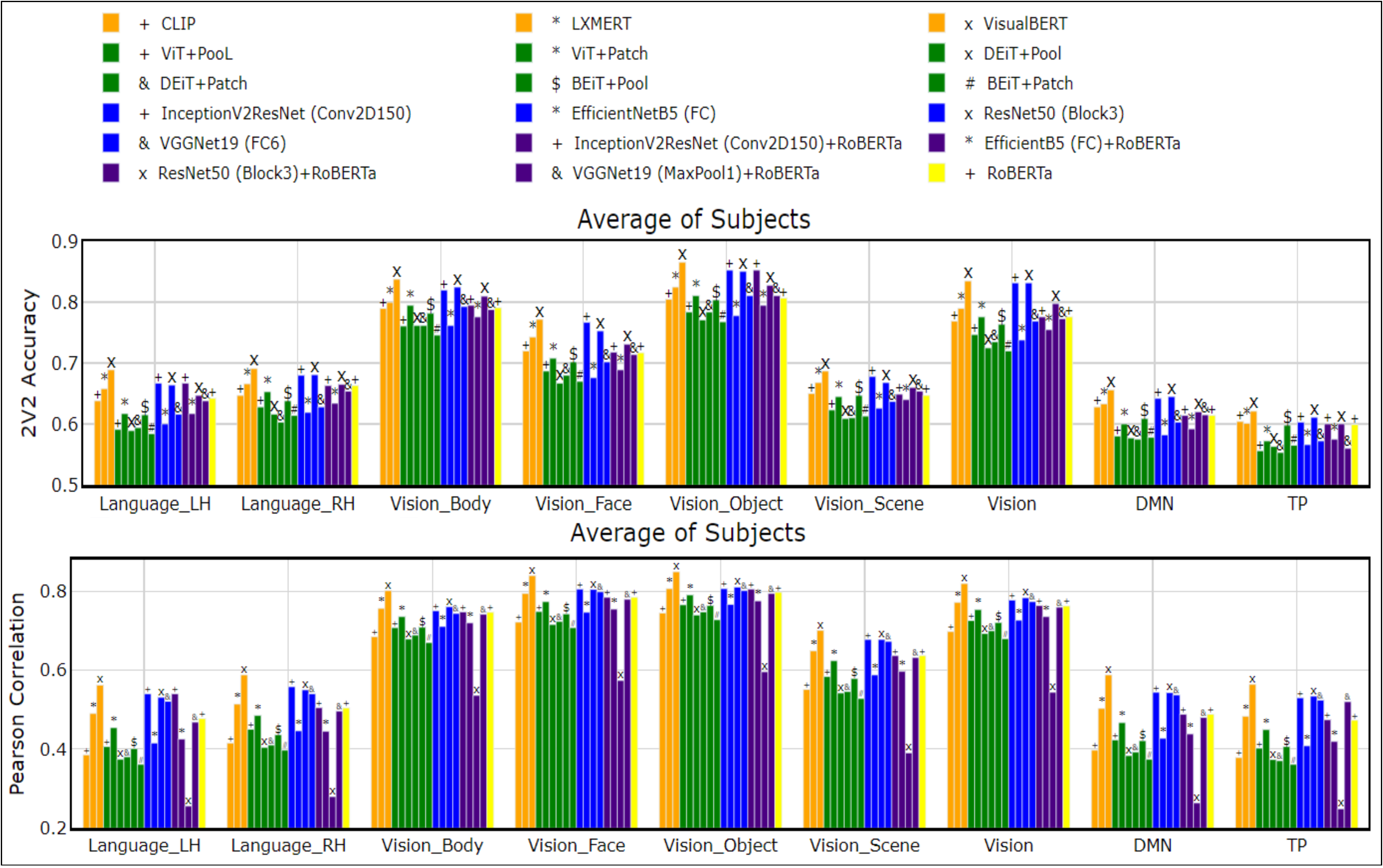}
\caption{Pereira Results: 2V2 (top figure) and Pearson correlation coefficient (bottom figure) between predicted and true responses across different brain regions using a variety of models. Results are averaged across all participants. VisualBERT performs the best.}
\label{fig:pereira_2v2_pcc_avg}
\end{figure*}

We present the 2V2 accuracy and Pearson correlation results for models trained with different input representations (features extracted from the best performing layer of every pretrained CNN model and last output layer of transformer model) on the two datasets: BOLD5000 and Pereira in Figs.~\ref{fig:Bold_2v2_pcc_avg} and~\ref{fig:pereira_2v2_pcc_avg}, respectively. We also compare the results using many intermediate layer activations (not just the best) for CNN models and last layer of Transformer models in Figs.~\ref{fig:bold_2v2_pcc_cnn_trans} and~\ref{fig:pereira_2v2_pcc_cnn_trans} in the Appendix. Further, we also compare the results using all intermediate layer activations for Transformer models in Figs.~\ref{fig:bold_pcc_trans}  and~\ref{fig:pereira_pcc_trans} in the Appendix.

\begin{figure*}[h] 
\centering
\includegraphics[width=0.9\linewidth]{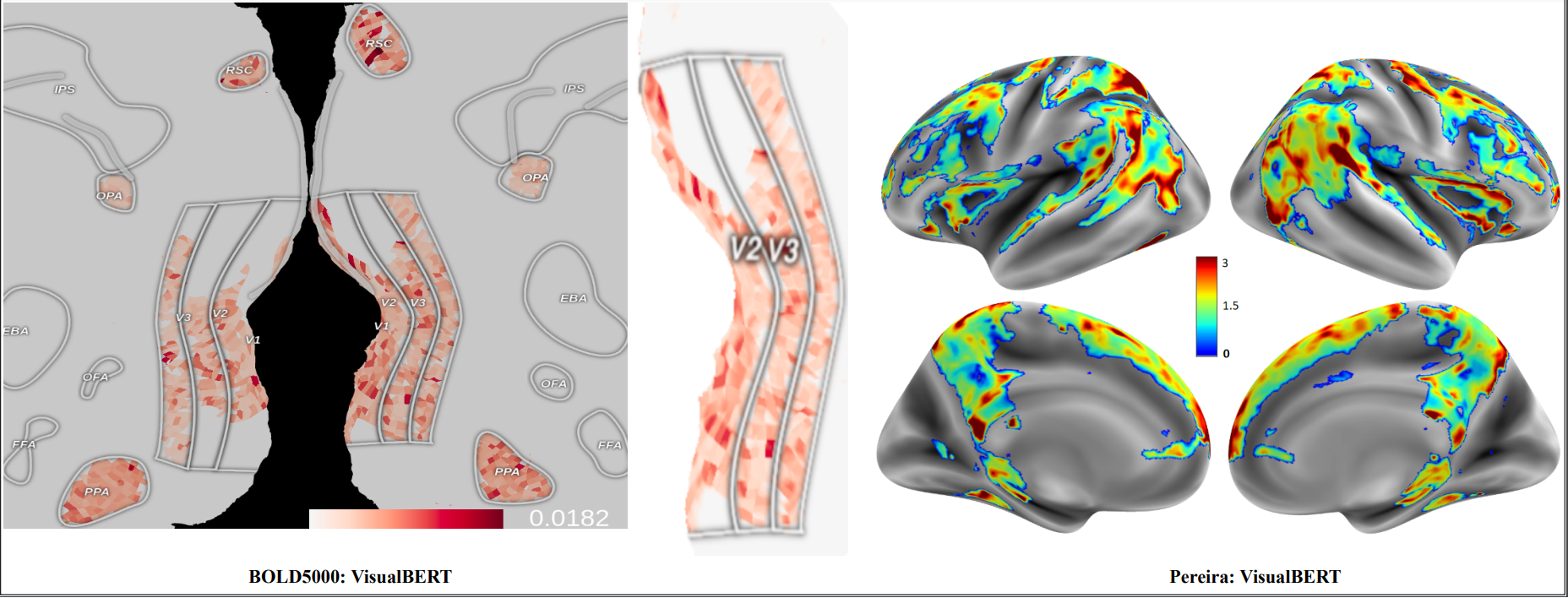}
\caption{MAE between actual and predicted voxels: (a) left figure is zoomed on V2 and V3 brain areas for VisualBERT on BOLD5000 subject 1. Note that V1 and V2 are also called EarlyVis area, while V3 is also called LOC area. (b) the right figure is for VisualBERT on the Pereira dataset subject 2.}
\label{fig:brainmaps_bold_pereira}
\end{figure*}

\noindent\textbf{BOLD5000:} We make the following observations from Fig.~\ref{fig:Bold_2v2_pcc_avg}: (1)
On both 2V2 accuracy and Pearson correlation, VisualBERT is better across all the models. (2) Other multi-modal Transformers such as LXMERT and CLIP perform as good as pretrained CNNs. Surprisingly, image Transformers perform worse than pretrained CNNs. Late fusion models and RoBERTa perform the worst. 
(3) Late visual areas such as OPA (scene-related) and LOC (object-related) display a higher Pearson correlation with multi-modal Transformers which is inline with the visual processing hierarchy. In general, a higher correlation with all the visual brain ROIs with multi-modal Transformers demonstrates the power of jointly encoding visual and language information. (4) The Patch representation of image Transformers shows an improved 2V2 accuracy and Pearson correlation compared to the Pooled representation. (5) Both InceptionV2ResNet and ResNet-50 have better performance among uni-modality models.

In order to estimate the statistical significance of the performance differences, we performed 2-tailed t-test for all the subjects across the five brain ROIs. We found that VisualBERT is statistically significantly better than LXMERT (second-best multi-modal Transformer) as well as InceptionV2ResNet (best pretrained CNN) for all ROIs except EarlyVis. Lastly, InceptionV2ResNet is statistically significantly better than BEiT (best image Transformer) for all ROIs. Detailed p-values are mentioned in Table~\ref{tab:pValBOLD} in the Appendix.


\noindent\textbf{Pereira:} We make the following observations from Fig.~\ref{fig:pereira_2v2_pcc_avg}: (1) Similar to BOLD5000, multi-modal Transformers such as VisualBERT and LXMERT perform better. (2) Lateral visual areas such as Vision\_Object, Vision\_Body, Vision\_Face, and Vision areas display higher correlation with multi-modal Transformers. A higher correlation with all the visual brain regions, Language regions, DMN, and TP with multi-modal Transformers, demonstrate that the alignment of visual-language understanding helps. 

In order to estimate the statistical significance of the performance differences, we performed 2-tailed t-test for all the subjects across the nine brain ROIs. We found that VisualBERT is statistically significantly better than LXMERT (second-best multi-modal Transformer) for all ROIs except Vision\_Body. Further, VisualBERT is statistically significantly better than ResNet (best pretrained CNN) for all ROIs except Vision\_Object and Vision\_Scene. Lastly, ResNet is statistically significantly better than ViT (best image Transformer) for all ROIs. Detailed p-values are mentioned in Table~\ref{tab:pValPereira} in the Appendix.

\begin{figure*}[h] 
\includegraphics[width=\linewidth]{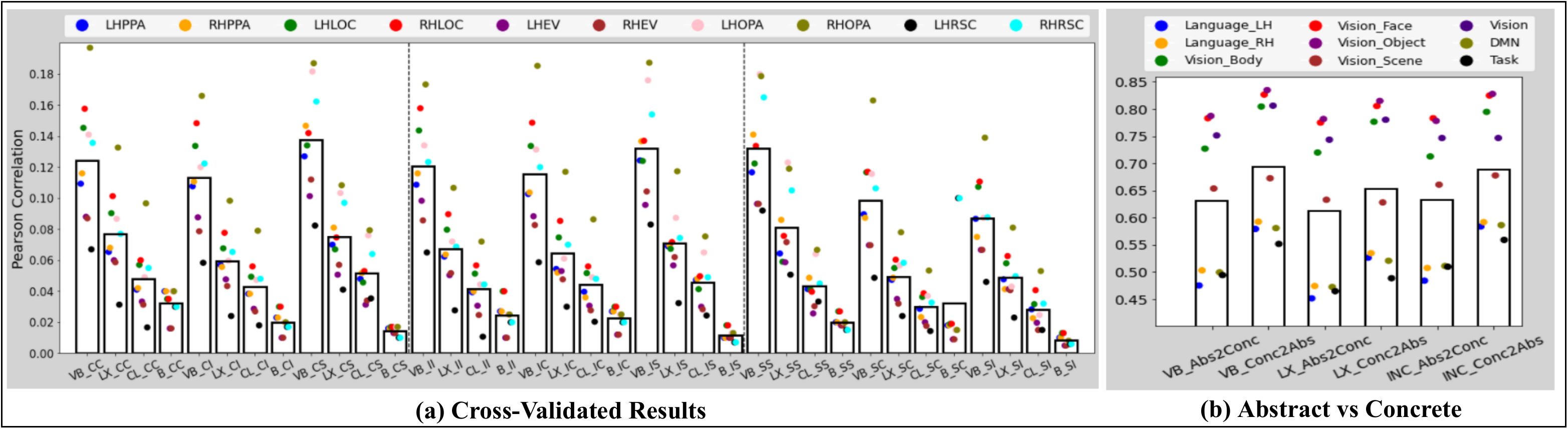}
\caption{(a) Cross-Validated Results for BOLD5000 dataset. (b) Abstract-Concrete Results for Pereira dataset. VB=VisualBERT, LX=LXMERT, CL=CLIP, B=Baseline~\citep{blauch2019assessing}, INC=InceptionV2ResNet. CC=Train and test on COCO, CI=Train on COCO and test on ImageNet, CS=Train on COCO and test on Scenes, and so on.)}
\label{tab:cross-validated_abscon}
\end{figure*}

As further analysis, in Fig.~\ref{fig:brainmaps_bold_pereira}, we show the mean absolute error (MAE) between the actual and predicted voxels across brain regions using VisualBERT. Comparing with similar brain charts for other models (shown in Figs.~\ref{fig:brainmaps_all_models} and~\ref{fig:brainmaps_pereira} in the Appendix), we notice that the error magnitudes are very small for the majority of the voxels. We observe that MAE values are relatively higher for EarlyVis areas and lowest for OPA, for BOLD5000.  
\subsection{Model size vs Efficacy Comparison}
We plot comparison of model size with Pearson Correlation (PC) averaged across all subjects for BOLD5000 in Fig.~\ref{fig:plot1}. We observe that compared to LXMERT, VisualBERT is not just more accurate but much smaller as well. VisualBERT is much more accurate compared to image Transformers while being of almost the same size. Lastly, pretrained CNNs are smaller compared to VisualBERT but are less accurate even when the particular layer activations used are cherry-picked. We observe similar trends for Pereira dataset as illustrated in Fig.~\ref{fig:plot2} in the Appendix.

\subsection{Cross-Validated fMRI Encoding}
Fig.~\ref{tab:cross-validated_abscon}(a) illustrates PC for cross-validated encoding on BOLD5000 using three multi-modal Transformers (VisualBERT, LXMERT, and CLIP). We also show results for a baseline method~\citep{blauch2019assessing}. 
We observe that (1) multi-modal Transformers outperform the baseline results across all the five brain regions for all the cross-validated tasks. (2) PC score is higher for train on COCO and test on ImageNet
in the object-selective visual area LOC (lateral occipital cortex), which makes sense since COCO has many objects. (3) Similarly, the scene-selective brain areas such as RSC and OPA have a higher correlation for the COCO-Scenes, ImageNet-Scenes, and Scenes-Scenes tasks. (4) EarlyVisual areas have a lower correlation compared to other brain regions across the three tasks. (5) Overall, the models trained on COCO or  ImageNet report higher correlation rather than those trained on Scenes.

\subsection{Abstract-Concrete fMRI Encoding}
The results for the abstract-train-concrete-test and concrete-train-abstract-test encoder models are presented across brain regions using two best multi-modal Transformers (VisualBERT and LXMERT), and the best pretrained CNN model (InceptionV2ResNet) in Fig.~\ref{tab:cross-validated_abscon}(b). We observe that the concrete-train-abstract-test model provides a better PC score compared to the abstract-train-concrete-test model. This matches our expectation that our brain can learn much better from concrete concepts than abstract concepts. PC analysis across brain regions provides the following insights. (1) The visual brain areas such as Vision\_Body, Vision\_Face, Vision\_Object, and Vision have superior performance for both concrete and abstract concepts; surprisingly, this is not the case for the Vision\_Scene area. (2) The language, DMN, and Task Positive (TP) brain networks have a higher correlation in the concrete-train-abstract-test than the abstract-train-concrete-test model. (3) Overall, VisualBERT and InceptionV2ResNet report similar performance with a slight edge over LXMERT.



\section{Cognitive Insights: Does Language Influence Vision?}
We discussed various insights in detail in Section~\ref{sec:experiments}. We summarize cognitive insights in the following. BOLD5000 dataset comprises brain responses from visual areas (early visual, scene-related, and object-related) when visual stimuli are presented to the subjects. Although only visual information is present in the stimuli, it is conceivable that participants implicitly invoke appropriate linguistic representations that in turn influence visual processing~\citep{lupyan2020effects}. Thus, it is not surprising that computational models such as multi-modal Transformers (VisualBERT, and LXMERT) that learn joint representation of language and vision show superior performance on the `purely' visual response data in BOLD5000 (see Figs.~\ref{fig:Bold_2v2_pcc_avg} and~\ref{fig:brainmaps_bold_pereira}(a)). 

Further, the performance of these models is naturally good in the case when text and image are shown to the participants, and whole brain responses are captured as in the case of the Pereira dataset (see Figs.~\ref{fig:pereira_2v2_pcc_avg} and~\ref{fig:brainmaps_bold_pereira}(b)). Based on the intuition from the computational experiments, we make the following testable prediction for future fMRI experiments. Instead of a passive viewing task, if participants were to perform a naming task/decision-making task on the objects/scenes, we expect to see more pronounced and focused activation in the visual areas 
during the language-based task as compared to passive viewing.

\section{Discussion}
Although VisualBERT seems to perform well, it is relatively large in size. Recently there has been a lot of work on compression of large deep learning models~\cite{cheng2017survey,deng2020model} which can be leveraged. Further, although we observe that VisualBERT leads to improved results, does it actually work brain-like? We plan to explore correlations between brain voxel space and representational feature space in the future to answer this questions. Finally, in this work we explored multimodal stimuli as a combination of vision and text. Combined strength of joint (audio, vision and text) modalities remains to be investigated.

\section{Conclusions}
In this paper, we studied the effectiveness of multi-modal modeling for brain encoding. We found that multi-modal visio-linguistic Transformers, which jointly encode text and visual input using cross-modal attention at multiple levels, perform the best. Our experiments on BOLD5000 and Pereira datasets lead to interesting cognitive insights. These insights indicate that fMRIs reveal reliable responses in scenes and object selection visual brain areas, which shows that cross-view translation tasks like image captioning or image tagging are practically possible with good accuracy. We plan to explore this as part of future work.

\section{Ethical Statement}
We reused publicly available datasets for this work: BOLD5000 and Pereira. We did not collect any new dataset. 

BOLD5000 dataset, except the stimulus images and their original annotations, is licensed under a Creative Commons 0 License. Please read their terms of use\footnote{\url{https://bold5000-dataset.github.io/website/terms.html}} for more details.

Pereira dataset can be downloaded from \url{https://osf.io/crwz7/}. Please read their terms of use\footnote{\url{https://github.com/CenterForOpenScience/cos.io/blob/master/TERMS_OF_USE.md}} for more details.

We do not foresee any harmful uses of this technology. 

\bibliography{references}
\bibliographystyle{acl_natbib}
\appendix

\setlength{\tabcolsep}{2pt}
\section{Dataset statistics}
We show number of instances and voxel distribution across various brain regions for the BOLD5000 and Pereira datasets in Tables~\ref{tab:bold5000_stats} and~\ref{tab:pereira_stats} respectively.

\begin{table}[!htb]
\scriptsize
\centering
\begin{tabular}{|l|c|c c |c c |c c |c c |c c|}
\hline
 & & \multicolumn{10}{c|}{Number of Voxels in Each ROI} \\ \cline{1-12} 
ROIs$\rightarrow$ & & \multicolumn{2}{c|}{PPA} & \multicolumn{2}{c|}{LOC} & \multicolumn{2}{c|}{EarlyVis} & \multicolumn{2}{c|}{OPA}  & \multicolumn{2}{c|}{RSC} \\ \cline{1-12} 
$\downarrow$Subjects& \#Instances& LH& RH & LH & RH& LH & RH & LH& RH & LH & RH 
\\ \hline 
Subject-1& 5254& 131&200&152&190&210&285&101&187&86&143 \\
Subject-2& 5254& 172&198&327&561&254&241&85&95&59&278  \\ 
Subject-3& 5254& 112&161&430&597&522&696&187&205&78&116 \\
Subject-4& 3108& 157&187&455&417&408&356&279&335&51&142\\
\hline
\end{tabular}
\caption{BOLD5000 Dataset Statistics. LH=Left Hemisphere. RH - Right Hemisphere.}
\label{tab:bold5000_stats}
\end{table}

\begin{table}[!htb]
\scriptsize
\centering
\begin{tabular}{|l|c c |c c c c c| c |c |}
\hline
 &\multicolumn{9}{c|}{Number of Voxels in Each ROI} \\ \cline{1-10}
ROIs$\rightarrow$ & \multicolumn{2}{c|}{Language} & \multicolumn{5}{c|}{Vision} & \multicolumn{1}{c|}{DMN} & \multicolumn{1}{c|}{Task Positive} \\ \cline{1-10} 
$\downarrow$Subj& LH& RH & Body & Face& Object & Scene & Vision& RH & LH  
\\ \hline 
P01&5265&6172&3774&4963&8085&4141&12829&17190&35120 \\
M01&5716&5561&3934&4246&7357&3606&12075&17000&34582 \\
M02&4930&5861&3873&4782&7552&3173&11729&15070&30594 \\
M03&3616&4247&2838&3459&5956&2822&9074&12555&24486 \\
M04&5906&5401&3867&4803&7812&3602&12278&18011&34024 \\
M05&4607&4837&2961&4023&6609&3135&10417&14096&28642 \\
M06&4993&5099&3424&4374&7300&4058&11986&16289&30109 \\
M07&5629&5001&4190&4993&8617&3721&12454&17020&30408 \\
M08&5083&5062&2624&4082&6463&3503&10439&14950&29972 \\
M09&3513&3650&2876&3343&5992&2815&9003&12469&25167 \\
M10&5458&5581&3232&4844&7445&3474&11530&16424&29400 \\
M13&4963&4811&2675&4008&5809&3323&9848&14489&30608 \\
M15&5315&6141&4112&4941&8323&3496&12383&15995&31610 \\
M16&4726&5534&4141&4669&8060&4142&12503&15104&31758 \\
M17&5854&5698&4416&4801&8831&4521&13829&16764&37463 \\
\hline
\end{tabular}
\caption{Pereira Dataset Statistics}
\label{tab:pereira_stats}
\end{table}





\section{Do multi-modal Transformers perform better encoding compared to intermediate layer representations from pretrained CNNs?}

\begin{figure*}[h]
\includegraphics[width=\linewidth]{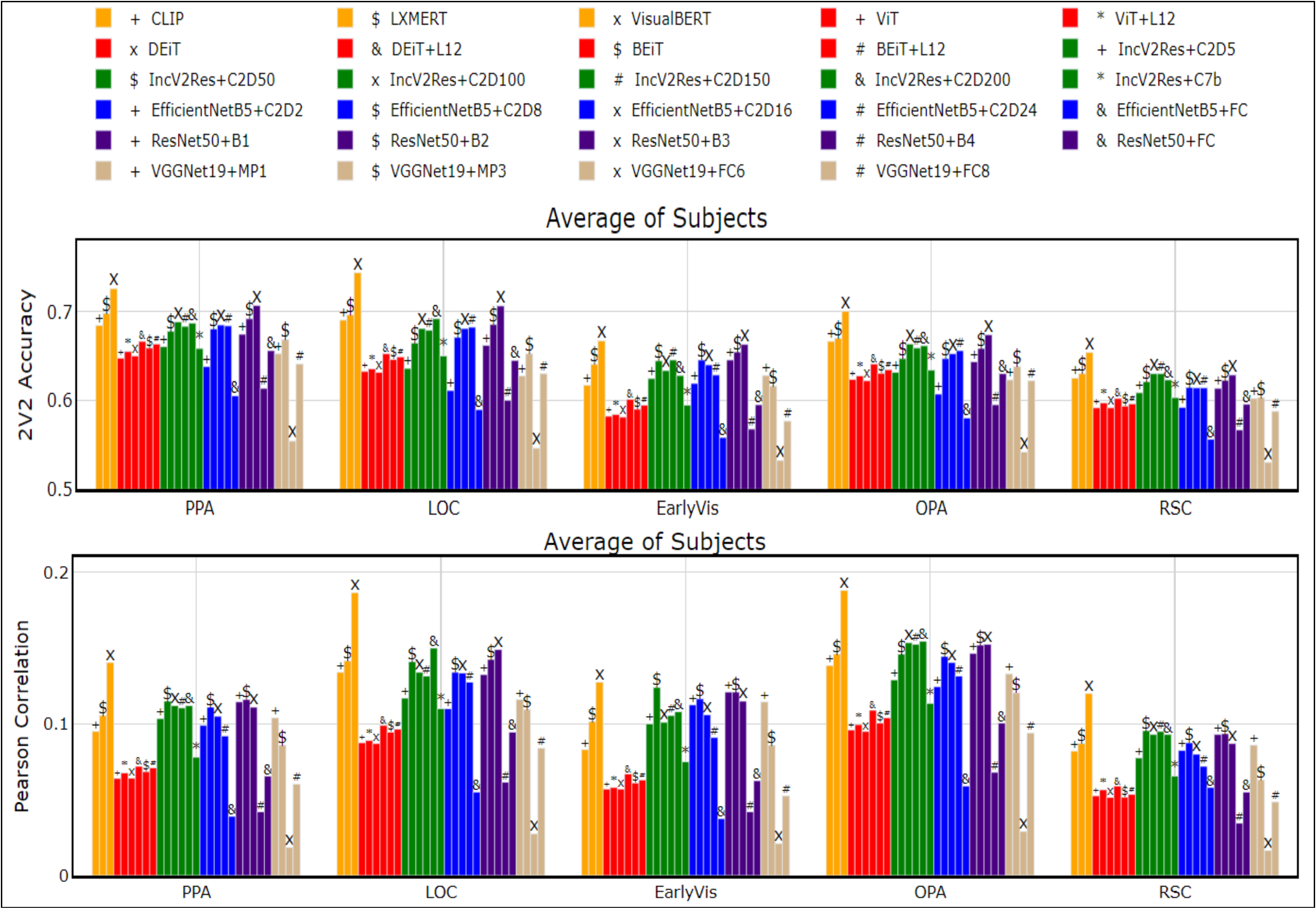}
\caption{BOLD5000: 2V2 (top Fig.) and Pearson correlation coefficient (bottom Fig.) between predicted and true responses across different brain regions using variety of models. Results are averaged across all participants. Pretrained CNN results are shown for all layers while multi-modal Transformer results are shown for last layers only.}
\label{fig:bold_2v2_pcc_cnn_trans}
\end{figure*}

\begin{figure*}[t]
\centering
\includegraphics[width=1.3\linewidth, angle =90]{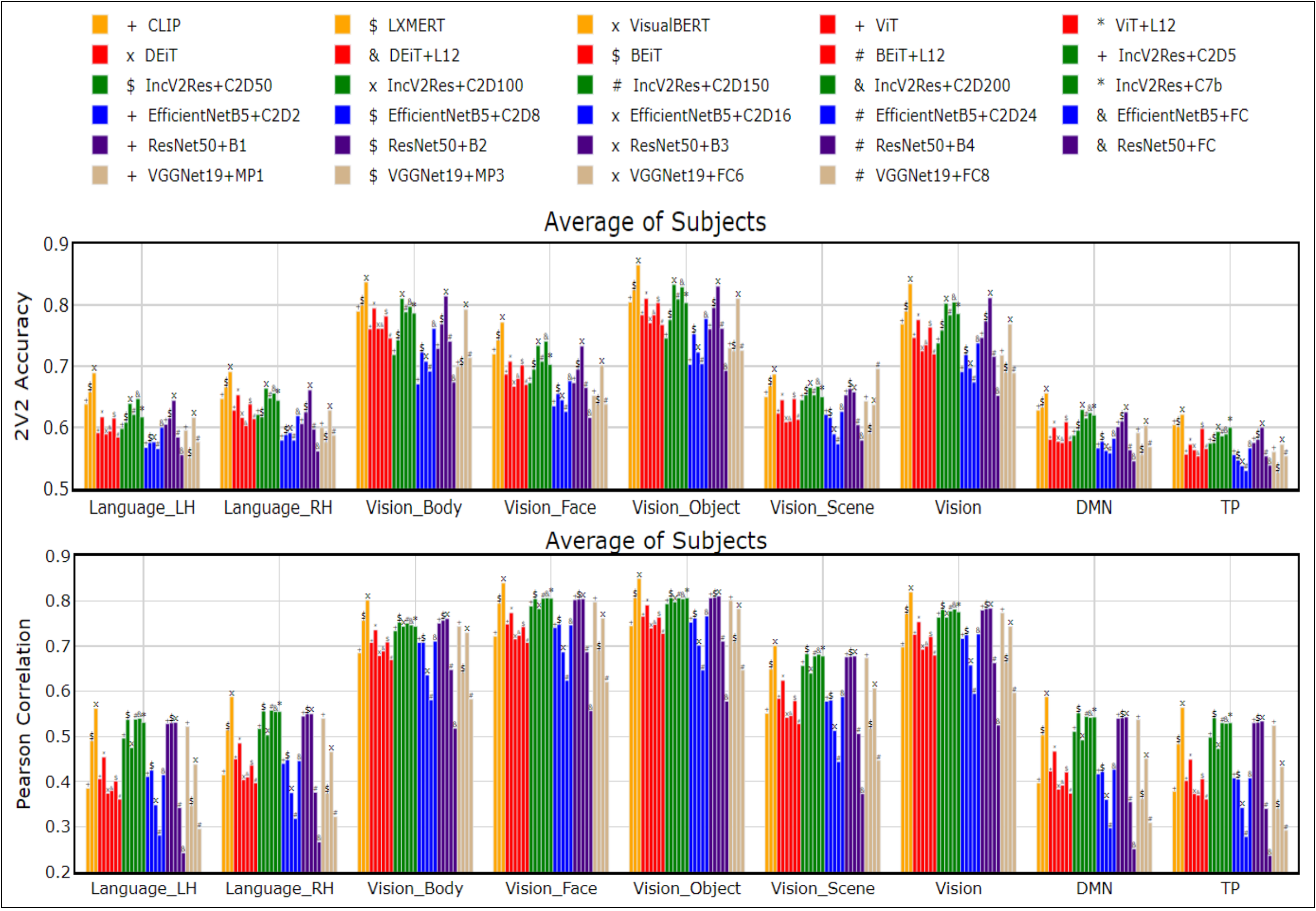}
\caption{Pereira dataset: 2V2 (top Fig.) and Pearson correlation coefficient (bottom Fig.) between predicted and true responses across different brain regions using variety of models. Results are averaged across all participants. Pretrained CNN results are shown for all layers while multi-modal Transformer results are shown for last layers only.}
\label{fig:pereira_2v2_pcc_cnn_trans}
\end{figure*}

We present the 2V2 accuracy and Pearson correlation for models trained with representations extracted from the last layer of multi-modal Transformers and all the lower to higher-level representations from pretrained CNNs on the two datasets: BOLD5000 and Pereira in Figs.~\ref{fig:bold_2v2_pcc_cnn_trans} and~\ref{fig:pereira_2v2_pcc_cnn_trans}, respectively.

\begin{figure}[t]
\centering
 \scriptsize 
\includegraphics[width=\linewidth]{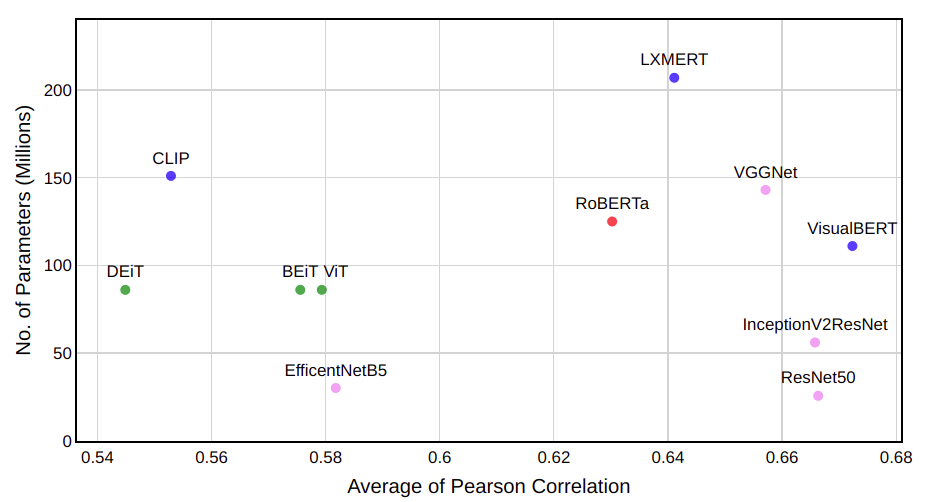}
\caption{Pereira: \#Parameters vs Avg Pearson Corr.}
\label{fig:plot2}
\end{figure}

We make the following observations from Fig.~\ref{fig:bold_2v2_pcc_cnn_trans}: (1) With respect to 2V2 and Pearson correlation, the multi-modal Transformer, VisualBERT, performs better than all the internal representations of pretrained CNNs. (2) In the pretrained CNNs, intermediate blocks have better correlation scores as compared to lower or higher level layer representations. (3) Other multi-modal Transformers, CLIP, and LXMERT, have marginal improvements over all the models except intermediate blocks such as Conv2D150 in InceptionV2ResNet.

We make the following observations from Fig.~\ref{fig:pereira_2v2_pcc_cnn_trans}: (1) With respect to 2V2 and Pearson correlation, the multi-modal Transformer, VisualBERT, performs better than all the internal representations of pretrained CNNs. (2) Similar to BOLD5000, the intermediate blocks have better correlation scores as compared to lower or higher level layer representations in the pretrained CNNs on Pereira Dataset. (3) Other multi-modal Transformer, LXMERT, have equal performance with intermediate blocks of each pretrained CNN model.

\begin{figure*}[t]
\centering
\includegraphics[width=0.9\linewidth]{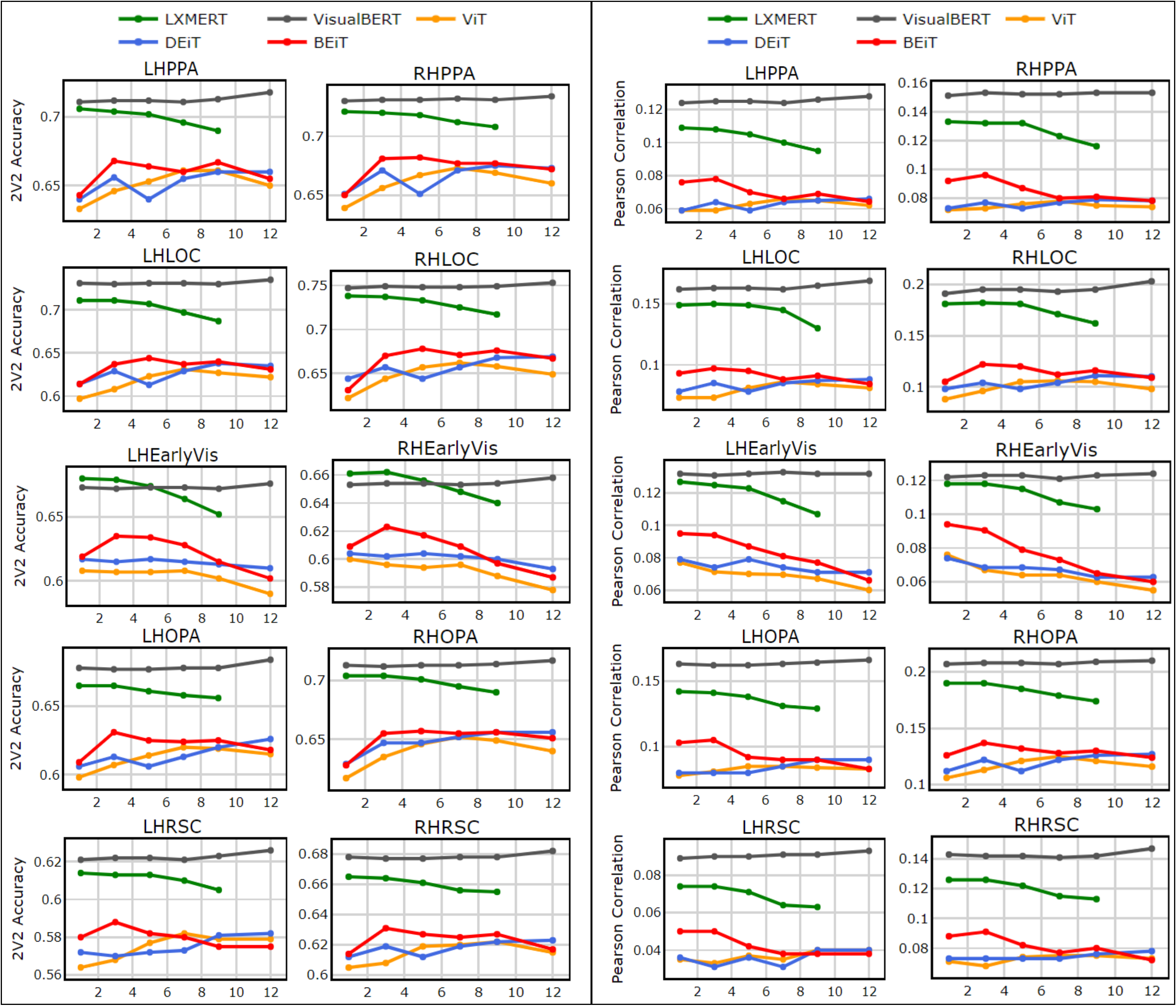}
\caption{BOLD5000: 2V2 (left) and Pearson correlation coefficient (right) between predicted and true responses across different brain regions using Transformer models. Results are averaged across all participants. The results are shown for all layers of image and multi-modal Transformers. Note that LXMERT has only 9 layers.}
\label{fig:bold_pcc_trans}
\end{figure*}

\begin{figure*}[t]
\centering
\includegraphics[width=0.9\linewidth]{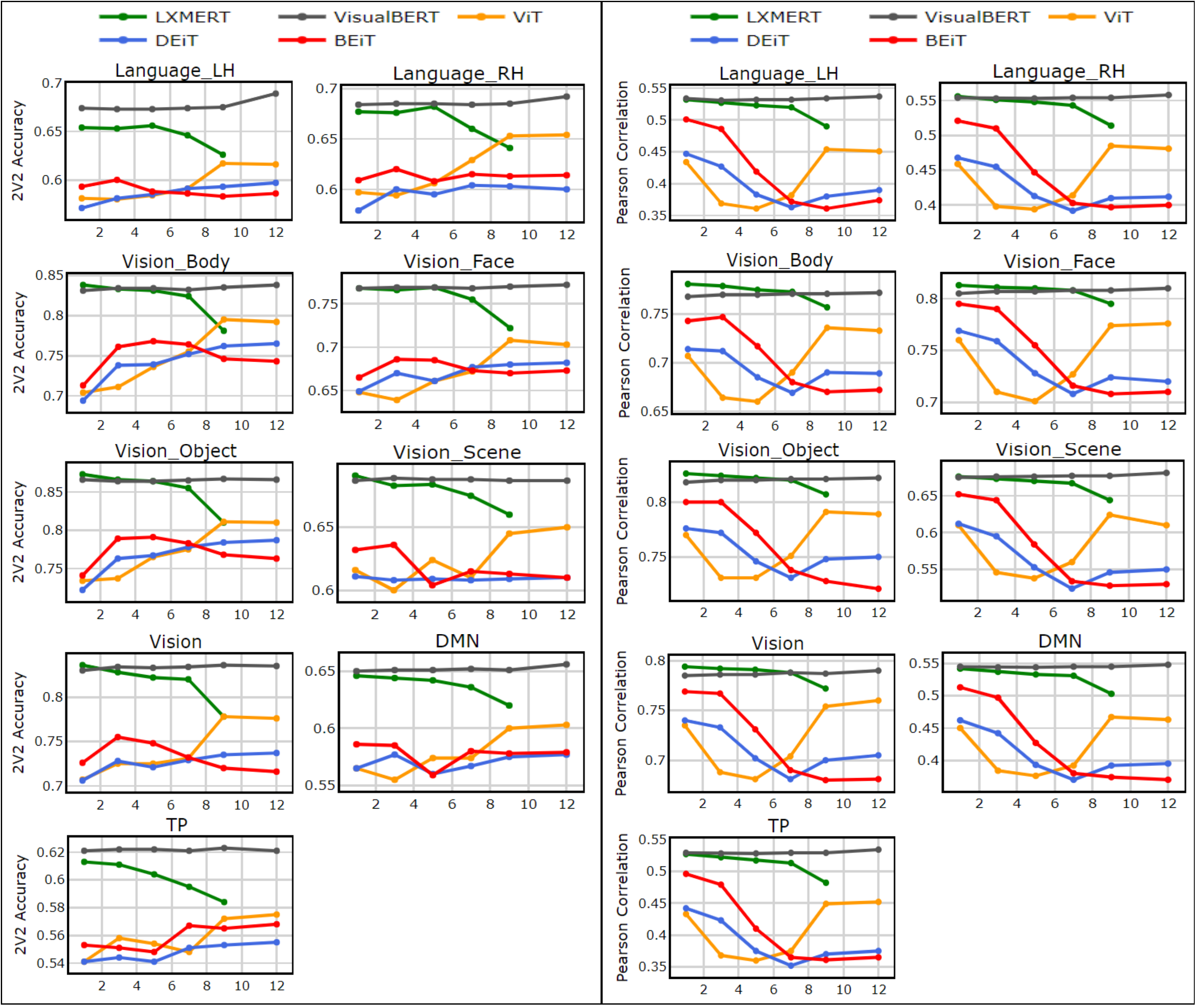}
\caption{Pereira: 2V2 (left) and Pearson correlation coefficient (right) between predicted and true responses across different brain regions using Transformer models. Results are averaged across all participants. The results are shown for all layers of image and multi-modal Transformers. Note that LXMERT has only 9 layers.}
\label{fig:pereira_pcc_trans}
\end{figure*}

\setlength{\tabcolsep}{3pt}
\begin{table}
    \centering
    \scriptsize
    \begin{tabular}{|p{1in}|c|c|c|c|c|}
    \hline
    Models compared &  PPA&LOC&EarlyVis&OPA&RSC\\
              \hline
                  \hline
    VisualBERT vs LXMERT&0.044*&0.004*&0.076&0.049*&0.029*\\
    \hline
    VisualBERT vs InceptionV2ResNet&0.049*&0.032*&0.521&0.041*&0.0354*\\
    \hline
InceptionV2ResNet vs BEiT &0.041*&0.003*&0.014*&0.188&0.203\\
\hline
    \end{tabular}
    \caption{p-values for 2-tailed t-test for BOLD5000 dataset}
    \label{tab:pValBOLD}
\end{table}

\setlength{\tabcolsep}{3pt}
\begin{table*}[!h]
    \centering
    \scriptsize
    \begin{tabular}{|p{1in}|c|c|c|c|c|c|c|c|c|}
    \hline
    Models compared &  Language\_LH&Language\_RH&Vision\_Body&Vision\_Face&Vision\_Object&Vision\_Scene&Vision&DMN&TP\\
    \hline
    \hline
    VisualBERT vs LXMERT&0.046*&  0.039*&  0.052&  0.048*&  0.046*&  0.045*&  0.047*&  0.040*&  0.035*\\
    \hline
    VisualBERT vs ResNet&0.049*& 0.038*& 0.048*& 0.048* &0.078&0.217& 0.048* & 0.046*&0.049*\\
    \hline
ResNet vs ViT &0.009* & 0.043*&0.041* & 0.047* &0.046* & 0.042* & 0.038* & 0.022* & 0.023*\\
\hline
    \end{tabular}
    \caption{p-values for 2-tailed t-test for Pereira dataset}
    \label{tab:pValPereira}
\end{table*}



\section{Do multi-modal Transformers perform better encoding in their layers?}
Given the hierarchical processing of visual or visual-language information across the Transformer layers, we further examine how these Transformer layers encode fMRI brain activity using image and mulit-modal Transformers. 
We present the layer-wise encoding performance results on two datasets: BOLD5000 and Pereira in Figs.~\ref{fig:bold_pcc_trans} and~\ref{fig:pereira_pcc_trans}, respectively.

\begin{figure*}[t] 
\includegraphics[width=\linewidth]{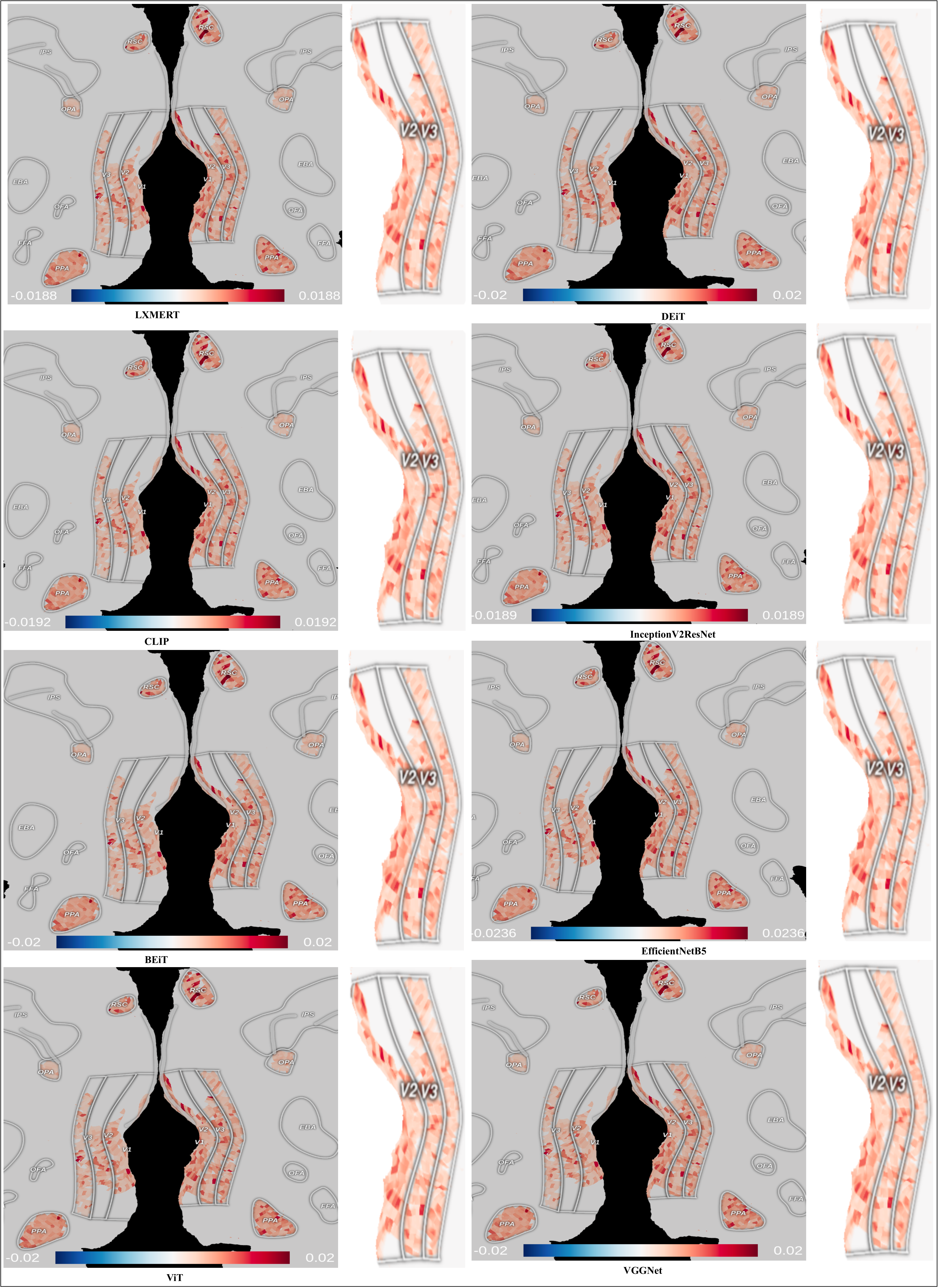}
\caption{MAE between actual and predicted voxels zoomed on V2 and V3 brain areas for various models. Note that V1 and V2 are also called EarlyVis area, while V3 is also called LOC area.}
\label{fig:brainmaps_all_models}
\end{figure*}

\begin{figure*}[h] 
\centering
\includegraphics[width=0.8\linewidth]{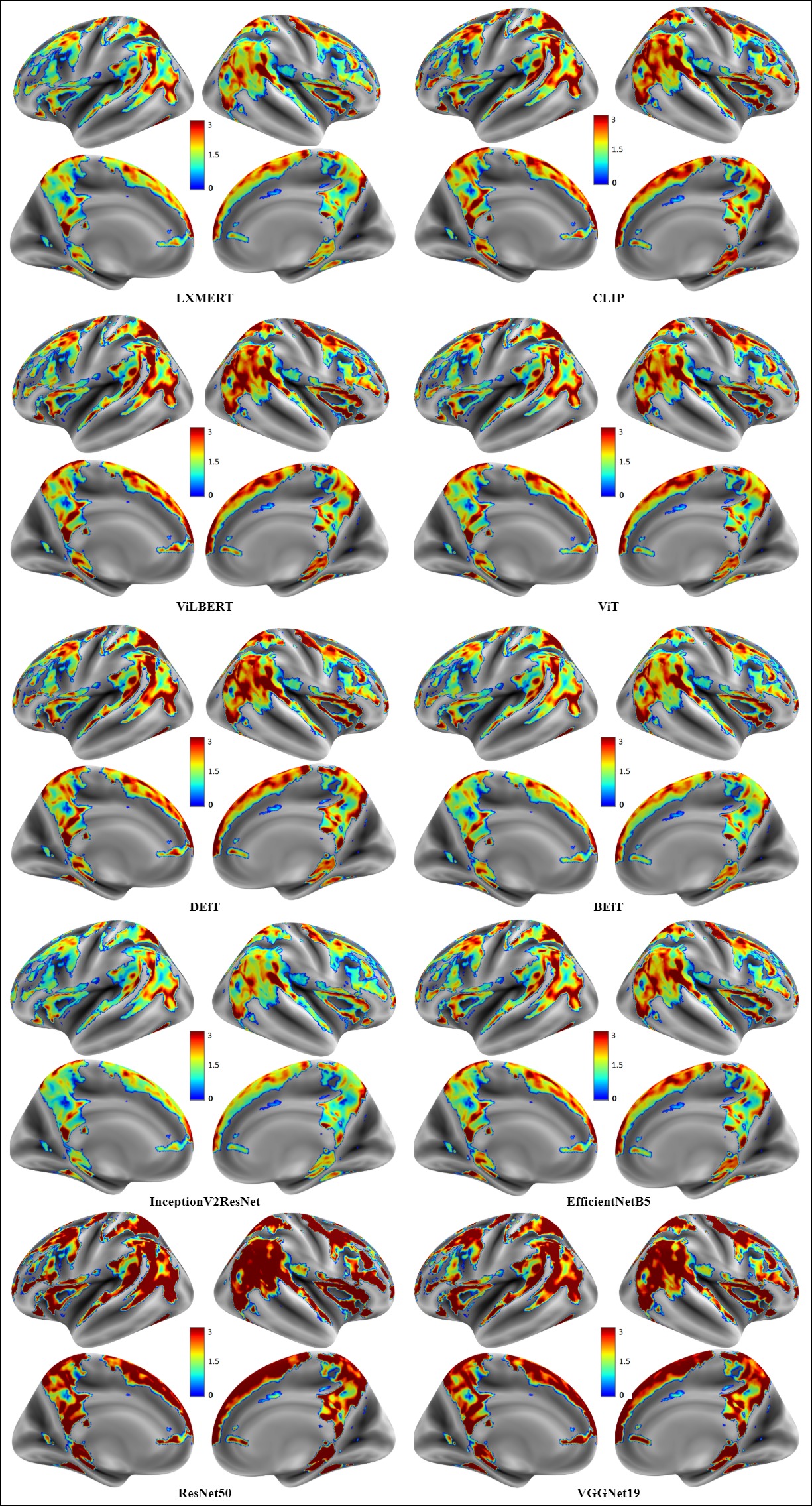}
\caption{MAE between actual and predicted voxels zoomed on V2 and V3 brain areas for various models. Note that V1 and V2 are also called EarlyVis area, while V3 is also called LOC area.}
\label{fig:brainmaps_pereira}
\end{figure*}

We make the following observations from Fig.~\ref{fig:bold_pcc_trans}: (i) The multi-modal Transformer, VisualBERT, have consistent performance across the layers from 1 to 12. (ii) The LXMERT model have marginal decreasing performance from intermediate layer (L7) to higher layers. (iii) The image Transformers have higher Pearson correlation for early visual areas in the lower layers whereas higher visual areas such as LOC, OPA, and PPA have an increasing correlation in higher layers. (iv) This clearly indicates that the hierarchy of processing of visual stimulus in the human brain is similar to image Transformer layers.

We make the following observations from Fig.~\ref{fig:pereira_pcc_trans}: (i) The multi-modal Transformers, VisualBERT, have consistent performance across the layers from 1 to 12. (ii) The LXMERT model have marginal decreasing performance from lower to higher layers. (iii) The image Transformer, ViT, has higher Pearson correlation for early visual areas in the lower layers whereas higher visual areas such as Vision\_Body, Vision\_Face, and Vision\_Obj have an increasing correlation in higher layers.

\section{Brain Maps for various models for BOLD5000 Dataset}

Fig.~\ref{fig:brainmaps_all_models} shows mean absolute errors (MAE) between actual and predicted voxels for various models on the BOLD5000 dataset. Notice that the magnitude of errors is much higher for a majority of voxels,  compared to that with the VisualBERT model as shown in Fig.~\ref{fig:brainmaps_bold_pereira}(a). Also, the multi-modal Transformers, VisaulBERT (MAE range: 0 to 0.0181) and LXMERT (MAE range: 0 to 0.0188), have lower MAE compared to both image Transformers (MAE range: 0 to 0.02) and pretrained CNNs (MAE range: 0 to 0.0236).

\section{Brain Maps for various models for Pereira Dataset}

Fig.~\ref{fig:brainmaps_pereira} shows mean absolute errors (MAE) between actual and predicted voxels for various models on the Pereira dataset. Notice that the magnitude of errors is much higher for a majority of voxels,  compared to that with the VisualBERT model as shown in Fig.~\ref{fig:brainmaps_pereira}(a). Also, the multi-modal Transformers, VisaulBERT and LXMERT, and InceptionV2ResNet+Conv2D150 have lower MAE compared to both image Transformers  and other pretrained CNNs.





\end{document}